\documentclass[10pt,twocolumn,letterpaper]{article}

\usepackage[pagenumbers]{wacv} % To force page numbers, e.g. for an arXiv version

\definecolor{wacvblue}{rgb}{0.21,0.49,0.74}
\usepackage[pagebackref,breaklinks,colorlinks,allcolors=wacvblue]{hyperref}

\def\wacvPaperID{1546} % *** Enter the WACV Paper ID here
\def\confName{WACV}
\def\confYear{2027}

\title{Spatial Support Matters: Geometry-Aware Graph Fusion for Rainfall Field Reconstruction}

\author{
\textbf{Low Jun Yu}$^{1,*}$\quad
\textbf{Niramay Kachhadiya}$^{2}$\quad
\textbf{Herath Mudiyanselage Viraj Vidura Herath}$^{2,+}$\\[0pt]
\textbf{Sanka Rasnayaka}$^{1}$\quad
and\ \textbf{Lucy Amanda Marshall}$^{2}$\\[2pt]
$^{1}$School of Computing, National University of Singapore, $^{2}$Faculty of Engineering, The University of Sydney\\
{\tt\small $^{*}$lowjy28@gmail.com, $^{+}$viraj.herath@sydney.edu.au}
}

\begin{document}
\maketitle
\begin{abstract}
Fine-scale rainfall reconstruction is critical for urban flood modeling, but real rainfall sensing systems observe the field through incompatible spatial supports: gauges measure points, microwave links measure paths, and radar/satellite products measure gridded areas. These differences in measurement support impose geometrically distinct constraints on the rainfall field, yet existing heterogeneous graph approaches reconcile such sources in feature space, giving each its own embedding while discarding the geometry of its support. We propose a geometry-aware multi-support heterogeneous graph neural network that represents each observation according to its support type (0D point, 1D line, or 2D grid) as a distinct node layer, and fuses them through cross-support message passing into a point-support prediction layer from which the field is reconstructed. An inductive masked-node formulation decouples prediction resolution from sensing resolution, allowing the same trained model to reconstruct the field at user-defined target locations or display grids. On Singapore data, the proposed method reduces RMSE by 23.2\% over the classical interpolation baseline, inverse-distance weighting, and consistently outperforms other neural architectures such as convolutional fusion and support-agnostic heterogeneous graph baselines. A generalization study using data from Sydney, Australia lets us characterize when multi-support fusion helps: the available skill appears to depend on gauge spacing relative to the spatial correlation length of the field, so fusion delivers the largest gains where the field is under-sampled relative to its correlation length and little when it is already resolved. Code and models are open-sourced at \url{<ANONYMIZED>}.

\end{abstract}
    
\section{Introduction}
\label{sec:intro}

Fine-scale rainfall reconstruction is critical for urban flood modeling, water management, and disaster response, yet rainfall remains one of the hardest geophysical fields to measure \cite{acosta2025physics}. Rainfall is spatially localized, intermittent, and rapidly evolving, with convective storms producing sharp gradients over a few kilometers that no single instrument captures well. Rain gauges sample it accurately but only at isolated points; weather radar and satellites cover wide areas but observe the field indirectly and aloft; commercial microwave links (CMLs) measure it along the paths between telecommunication towers. Each instrument senses the same field through a different spatial geometry, and no single source is sufficient alone. 

The difficulty is not only that the sensors are heterogeneous, but that they observe rainfall over different spatial supports: gauges measure point values, microwave links measure path-integrated attenuation, and radar or satellite products estimate areal averages over grid cells. We refer to this geometric measurement region as the \emph{spatial support} of an observation. These differences in support are not identical: a point measurement, a path integral, and an areal average impose geometrically distinct constraints on the field, and fusing them well requires respecting that geometry. 

%Heterogeneous graph neural networks (HGNNs) are a natural tool for fusing multi-source observations, as they can represent different sources as distinct node and edge types \citep{HGNNSurvey}. However, existing heterogeneous formulations treat the differences between sources as differences in \emph{feature space}: each source is given its own learned embedding, and the geometric meaning of its support is discarded. A point observation and a gridded observation are reconciled as two feature vectors attached to nodes, rather than as two different kinds of spatial constraint on a shared field, so the structural distinction between zero-, one-, and two-dimensional (0D, 1D, 2D) supports is never represented in the graph. The closest prior work \citep{yang2025offgrid} fuses point and gridded sources in a heterogeneous graph but models their difference only in feature space. It does not address line-integrated observations, and demonstrates prediction only at fixed observation locations rather than reconstructing a field at unobserved points.
%Existing geostatistical approaches such as inverse-distance weighting (IDW) and Krigging rest on stationary and Gaussianity assumptions leading to violation of non-linearities in convective rainfalls. Existing fusion approaches, including convolutional models and heterogeneous graphs, typically reconcile sensors as feature channels or node types. They do not explicitly model whether a measurement constrains a point, a line, or an area of the latent rainfall field.
Classical interpolation techniques such as inverse-distance weighting (IDW) and kriging \citep{goovaerts2000, Merging_review} rely on assumptions of stationarity and Gaussian spatial statistics, which are often violated by the highly non-linear, localized nature of convective rainfall. More recent learning-based approaches, including convolutional networks \cite{CNNDataFusion, wu2020deepfusion} and heterogeneous graph neural networks \cite{HGNNSurvey, CNGAT}, relax these assumptions by learning non-linear relationships directly from data and by integrating multiple sensing modalities. However, these methods typically reconcile heterogeneous sensors as feature channels or node types, without explicitly modeling the spatial support of each observation—whether it constrains the rainfall field at a point, along a line, or over an area.

%We propose a geometry-aware multi-support HGNN that represents each observation according to its support geometry. We evaluate the framework on rainfall estimation, which is among the few problems where all three support types are physically available: point rain gauges, line-integrated CML, and gridded radar and satellite products all observe the same precipitation field. Prior fusion approaches in this domain have relied on learned spatial interpolation \citep{KrigingCN, KrigingCN2} or convolutional models over gridded inputs \citep{CNNDataFusion, AdvancesInMSDF}, none of which represent the differing spatial supports of the sources within a single geometry-aware model. Rainfall thus serves as a demanding testbed for multi-support fusion, while the framework itself is general. 
We propose a geometry-aware multi-support Heterogeneous Graph Neural Network (HGNN) for rainfall-field reconstruction. The system represents gauges, CMLs, and gridded remote-sensing products according to their measurement support, and fuses them through cross-support message passing into a point-support prediction layer. This allows rainfall to be queried at unobserved locations while preserving the geometry by which each sensor constrains the field.

On the real-world Singapore testbed, the proposed support-aware model reduces root mean square error (RMSE) by 23.2\% over IDW and outperforms convolutional fusion and support-agnostic graph baselines.

Our contributions are: 
\begin{itemize}
    \item A practical multi-support rainfall reconstruction system that fuses point rain gauges, line-supported CMLs, and gridded radar/satellite products in a single geometry-aware graph.
    
    \item A support-aware graph construction that represents 0D point, 1D line, and 2D gridded observations as distinct support-typed layers and routes information through cross-support message passing.
    
    \item An inductive masked-node prediction protocol that prevents gauge-value leakage and enables rainfall estimates at unobserved target locations and user-defined output grids.
    
    \item A two-testbed application evaluation across Singapore and Sydney, showing that explicit support modeling improves reconstruction in under-sampled rainfall regimes and characterizing when multi-support fusion provides limited additional gain.
\end{itemize}

\section{Related Work}

%-------------------------------------------------------------------------
\subsection{Multi-Source Spatial Field Estimation}
Estimating a continuous rainfall field from sparse, heterogeneous sensors is a long-standing problem. Classical approaches include deterministic interpolation, geostatistical interpolation, radar bias adjustment, and radar-gauge integration through methods such as mean-field bias adjustment, kriging with external drift, and Bayesian merging among the most widely used techniques \citep{Merging_review}. Variational schemes that jointly retrieve rainfall from gauges, radar and microwave links have likewise shown that combining sources of differing character improves the resulting estimate \citep{Bianchi}. Many of these methods rely on stationarity, smoothness, or Gaussian spatial-statistical assumptions that can be violated by localized convective rainfall, and they tend to degrade in precisely the sparse, localized regimes where fusion is most valuable \citep{Merging_review}. 

Learned methods relax these assumptions by fitting non-linear mappings directly from data. Random-forest regression and convolutional networks have been applied to multi-source precipitation fusion \citep{rfmep2020, AdvancesInMSDF, CNNDataFusion}, with random-forest merging combining gauges, gridded products, and topography in data-scarce regions \citep{rfmep2020}, and convolutional fusion of radar, satellite, and gauges yielding the largest gains where gauge coverage is sparse \citep{CNNDataFusion}. Within graph learning, Graph Neural Networks (GNNs) have been applied to radar-based quantitative precipitation estimation \citep{CNGAT}, but predominantly on homogeneous graphs in which all sensors are treated as equivalent nodes. Across this progression of geostatistical, convolutional, and graph-based methods, sensor heterogeneity is handled in \emph{feature space}: each source contributes a different channel or node-feature vector, while the geometry of the measurement itself is left unmodelled. A gauge, a microwave link and a radar/satellite pixel are not interchangeable point samples; they constrain the underlying field over a point, a line and an area respectively, and the existing fusion methods do not represent this distinction in measurement support.

\subsection{Heterogeneous and Off-Grid Graph Learning}
HGNNs generalise message passing to graphs with multiple node and edge types $G = (V,E, \tau, \phi)$, where the type maps $\tau$ and $\phi$ enable type-specific transformations and metapath-based relational semantics \citep{HGNNSurvey}. GNNs have proven effective for large-scale geospatial modeling, including global weather emulation \citep{lam2023graphcastlearningskillfulmediumrange}. Standard HGNNs, however, distinguish sensors by node or edge \emph{type} (which variables a node carries) without modeling the spatial support over which each sensor constrains the field. Heterogeneity of feature space is represented; heterogeneity of measurement geometry is not. Inductive graph formulations have been developed for spatiotemporal kriging, recovering signals at unsampled locations by masking nodes and reconstructing them on sampled subgraphs \citep{ignnk2021}; these operate on homogeneous point sensors, with no notion of differing measurement support.

The closest prior work to ours is \citet{yang2025offgrid}, who fuse off-grid weather stations with gridded numerical-weather products on a heterogeneous graph for localized weather forecasting. Station nodes and grid-cell nodes form two node types, and each station aggregates information from neighboring stations and grid cells through message passing. This is structurally close to the gauge-plus-grid component of our framework, and the nearest point of comparison in the literature. We differ on three axes. First, both of their node types are treated as point-supported in the graph: weather stations and grid cells are represented by their coordinates and connected by Euclidean nearest-neighbor edges. Their graph thus encodes heterogeneity of feature space but not of spatial support. Our framework introduces line-supported observations as a first-class support type, with edges constructed from the geometry of the link rather than from point-to-point distance. Second, \citet{yang2025offgrid} predict at the fixed locations of known training stations under a temporal split, and do not reconstruct the field at unobserved locations; whereas our inductive setting evaluates spatially held-out locations and allows unseen target nodes to be inserted at inference on user-defined target grids. Third, their task is correction of an existing gridded forecast toward local observations, whereas ours is direct estimation of a spatial field from concurrent multi-support observations. We position our contribution as the multi-support abstraction these distinctions imply: a single graph in which point, line and gridded supports coexist and exchange messages, with measurement geometry preserved in the graph structure itself. 

\subsection{Line-Supported Observations}

CMLs provide the canonical line-supported measurement. Each link measures the path-averaged attenuation of the signal between two telecommunication towers, yielding a near-surface, path-integrated rainfall estimate along its span \citep{Rainlink}. In dense urban networks (Singapore among them), large numbers of links are already deployed by mobile operators, providing rainfall-relevant observations where gauges are sparse, at little additional cost \citep{CMLReview}. CML observations are often converted through multi-step pipelines before fusion; each step can introduce uncertainty that propagates into downstream estimates \citep{Rainlink}; coverage is also uneven, with gaps over water bodies, terrain, and restricted airspace \citep{CMLReview}. These properties make CMLs both valuable and awkward for existing fusion methods: their line geometry does not fit the point-node assumption underlying prior graph-based approaches, and prior work typically reduces links to representative points or pre-converted rainfall values before fusion. In our framework, a CML is treated as a single line-supported observation parameterized by its two endpoints, with cross-support edges constructed from line geometry rather than from a collapsed midpoint representation.

\section{Methodology}
\label{sec:method}
We consider the problem of inferring a continuous two-dimensional physical field from a heterogeneous collection of observations that measure the same underlying process through different \emph{spatial supports}. The spatial support of an observation is the geometric region over which it constrains the latent field. We distinguish three support types. A point (0D) observation constrains the field at a single location. A line (1D) observation constrains a path integral of the field along a one-dimensional segment. A gridded (2D) observation constrains an areal average of the field over a regular cell. Observations of different support are complementary, but cannot be optimally fused by treating them as generic graph nodes that differ only in their feature vectors. This discards how each observation geometrically constrains the field.

We propose a geometry-aware multi-support HGNN that preserves these differences. The framework represents observations as nodes organized into support-typed layers, encodes each according to its measurement geometry, and fuses them via cross-support message passing into a designated point-support prediction layer, from which the field is decoded. Let $\mathcal{G}=(\mathcal{V},\mathcal{E})$ denote the heterogeneous graph. The node set $\mathcal{V} = \mathcal{V}_{0} \cup \mathcal{V}_{1} \cup \mathcal{V}_2$ is partitioned by spatial support into point, line, and gridded nodes. In the rainfall application, $V_0$ contains rain gauges and target query locations, $V_1$ contains CML line supports, and $V_2$ contains radar or satellite grid cells. The edge set $\mathcal{E}$ is partitioned into within-support edges, which connect observation nodes that share a support type, and cross-support edges, which connect observation nodes to the prediction layer. Each support layer may itself contain multiple heterogeneous sources of the same dimensionality: two distinct gridded sources, for instance, each become a separate node type within the 2D support layer, each with their own learned message-passing parameters, but constructed using the same geometric procedure for cross-support edges. The framework is therefore modular: adding or removing a source requires only instantiating or dropping the corresponding node type and its cross-support edges, leaving the rest of the architecture unchanged. We exploit this in Section~\ref{sec:experiments}, where the same framework is instantiated on two testbeds with different support compositions.  

\subsection{Support-Typed Node Layers}
\label{sec:nodes}

Nodes in $\mathcal{G}$ are organised into three support layers, each encoding observations according to their measurement geometry. 

\textbf{Point (0D) nodes.}
Each point-support observation is represented as a single node $v \in \mathcal{V}_0$ at its geographical coordinate $x_v \in \mathbb{R}^2$. The 0D layer serves a dual role: observed nodes carry measurements of the latent field at known locations, while target nodes have their observations masked to zero and are the locations at which the field is to be predicted. Both are structurally identical 0D nodes: the masking is applied to node features at training time, not to the graph topology, so the target values must be inferred entirely from messages propagated by neighboring observed and cross-support nodes.

\textbf{Line (1D) nodes.} 
Each line-support sensor is represented as a pair of nodes $\{v_A, v_B\} \subset \mathcal{V}_1$ corresponding to its two physical endpoints.
In the graph implementation, the line support is encoded by endpoint nodes tied by a shared measurement and a within-support edge. This preserves the line-support geometry explicitly; the sensor is not collapsed to a single proxy point such as its midpoint, but retained as two endpoint nodes spanning the full measurement path.

\textbf{Gridded (2D) nodes.} 
Each cell of a gridded field product is represented as a single node $v \in \mathcal{V}_2$ positioned at the cell centre. Where multiple gridded sources are present, each source occupies a distinct node type with its own learned message-passing parameters, but is constructed using the same geometric procedure for cross-support edges described in Section~\ref{sec:edges}.

\textbf{Node features.} 
Point-support (0D) nodes carry the observed value together with a validity flag that distinguishes a genuine zero observation from a masked or missing one, and a Laplacian positional encoding (LPE) given by the four smallest non-trivial eigenvectors of the graph Laplacian \citep{dwivedi2021graphtransformer}. The validity flag is necessary because a masked target node and a true dry reading both present a value of zero; encoding observation validity as an explicit feature follows prior work on learning from sparse and partially-observed spatiotemporal graphs \citep{marisca2022}. The LPE assigns each node a structurally unique position in the graph, letting the model distinguish nodes that carry similar observed values but occupy different positions in the network. For line-supported CMLs, both endpoint nodes share the same link measurement, with the endpoints defining the path geometry rather than two independent point observations. Gridded nodes carry the corresponding radar or satellite cell value.

\subsection{Edge Construction}
\label{sec:edges}
\begin{figure}[!htb]
    \centering
    \includegraphics[width=1.0\linewidth]{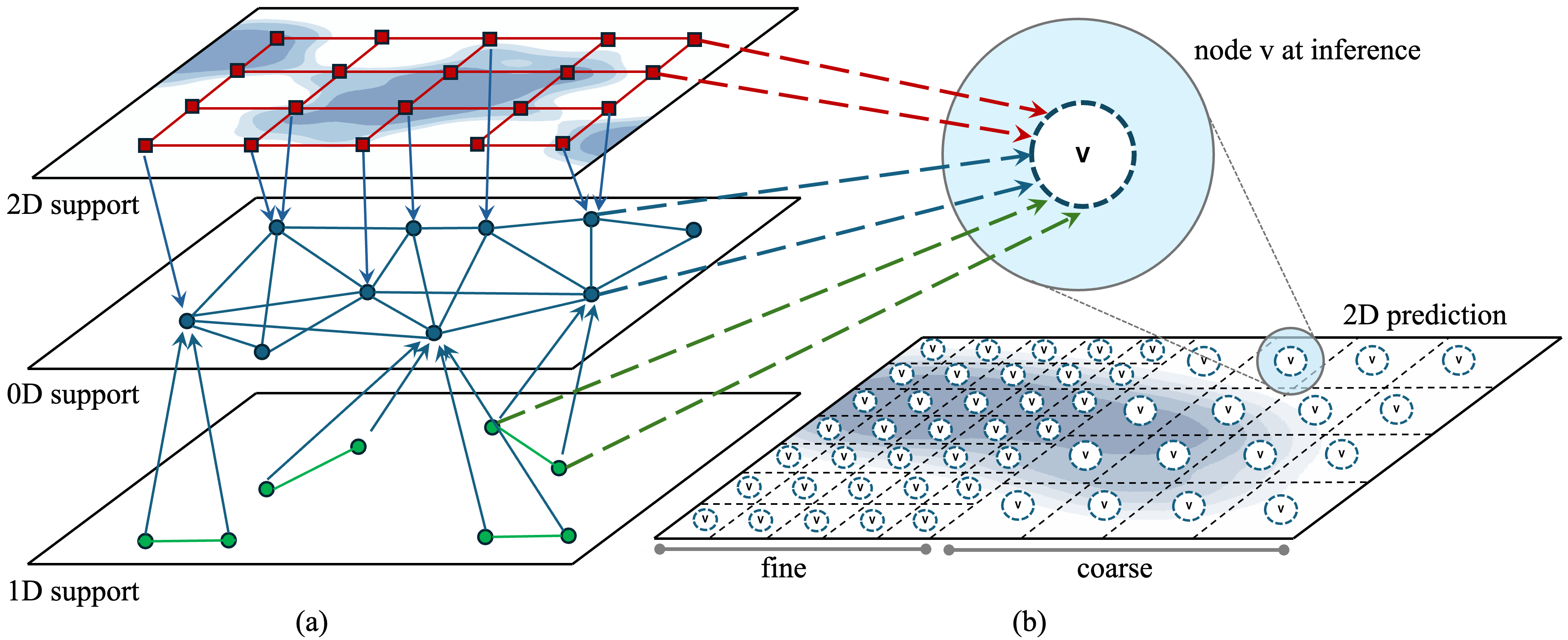}
    \caption{(a) Heterogeneous graph construction over three measurement support types: 2D gridded nodes (red squares), 1D line nodes (green circles), and 0D point nodes (dark blue circles). Cross-support edges route information from 2D and 1D nodes into the 0D prediction layer. (b) At inference, a virtual node v is inserted at any target location and connected to the existing graph via cross-support edges (zoom inset). Repeating this process across a regular grid yields a dense predicted field at a user-defined display resolution (fine, left; coarse, right).}
    \label{fig:edges figure}
\end{figure}

Edges connect nodes either within a single support layer or across support layers into the prediction layer (Figure~\ref{fig:edges figure}). Spatial proximity edges are weighted by normalized inverse distance. The internal edge connecting the two endpoints of a CML uses the physical path length as its line-support attribute. For a pair of nodes $i,j$ with Euclidean distance $d_{ij}$, the weight is normalised against the strongest connection at the target node, 
\begin{equation}
    w_{ij} = \frac{1 / d_{ij}}{\max_{k \in \mathcal{N}(j)} 1 / d_{kj}}, 
    \label{eq:weight}
\end{equation}
so that weights lie in $(0,1]$ and the nearest neighbour of each target node receives unit weight. This normalisation is applied to all edge types described below.

\textbf{Within-support edges.}
Point (0D) nodes are connected to their $k$ nearest point neighbors using a ball-tree search over node coordinates, with weights given by \cref{eq:weight}. Each line (1D) sensor contributes a single within-support edge joining its two endpoint nodes, weighted by physical length of the integration path. Gridded (2D) nodes are connected to their immediate neighbors on the regular grid, each cell linking to its surrounding cells in its local (eight-neighbor) neighborhood, so that within-support message passing propagates information across the gridded field.

\textbf{Cross-support edges.}
Every observation layer is connected to the point-support prediction layer through cross-support edges, and these are the only paths by which line and gridded observations reach the prediction. For a gridded source, each prediction node is connected to its $k$ nearest gridded nodes under the haversine metric, weighted by \cref{eq:weight}. 

For a line source, connectivity respects the line geometry of the sensor rather than approximating it by a point. For a prediction node $v$ and a line sensor with segment geometry $\ell$ (where $p$ denotes any point along the link path), the selection distance is the perpendicular distance from the node to the segment, 
\begin{equation}
    d_{\perp}(v,\ell) = \min_{p \in \ell} \, \lVert x_v - p \rVert_2,
    \label{eq:perp}
\end{equation}
and the $k$ line sensors minimising \cref{eq:perp} are selected for each prediction node. The node is then connected to \emph{both} endpoint nodes of each selected sensor, with each endpoint edge weighted by \cref{eq:weight} using its own node-to-endpoint distance. A prediction node therefore receives a separate, distance-appropriate message from each end of the integration path, preserving the line support of the observation rather than collapsing it to the midpoint.

\textbf{Leakage Control.}
Cross-support edges are directed into the prediction layer, and all edges are reconstructed independently for each data split. Target nodes held out for evaluation never enter the training-graph topology, so no information about their location or connectivity is available at training time.

\subsection{Message Passing}
\label{sec:mp}

Given the constructed graph, the prediction at each point-support node is computed by message passing over the heterogeneous edge set. We use a heterogeneous convolution in which a separate graph convolution is instantiated for every edge type, so that messages arriving along different support relations are transformed by their own learned parameters before being combined, implemented with the heterogeneous convolution wrapper of PyTorch Geometric \citep{pyg2019}. For a target node $v$ at layer $l$, each incoming edge type $r$ produces an aggregated message, and the messages from all edge types incident on $v$ are combined by mean aggregation:

\begin{equation}
    h_v^{(l+1)} = \sigma\!\Bigg(
    \underset{r \in \mathcal{R}(v)}{\mathrm{mean}} \;
    \sum_{j \in \mathcal{N}_r(v)} w_{jv} \, h_j^{(l)} \mathbf{W}_r^{(l)}
    \Bigg), 
    \label{eq:mp}
\end{equation}
where $\mathcal{R}(v)$ is the set of edge types incident on $v$,
$\mathcal{N}_r(v)$ is the neighbourhood of $v$ under edge type $r$, 
$w_{jv}$ is the inverse-distance edge weight from \cref{eq:weight}, 
$\mathbf{W}_r^{(l)}$ is the learned weight matrix for edge type $r$ at layer $l$, and $\sigma$ is a non-linear activation. A shared hidden dimension is used across all node types so that embeddings from different support layers are directly comparable when aggregated at a prediction node.

Stacking $L$ such layers allows each prediction node to aggregate information from progressively larger neighbourhoods: cross-support edges admit line and gridded observations into the point-support layer, and within-support edges then propagate this fused information between point nodes. Each hidden layer applies a ReLU non-linearity, while the final linear layer maps the embedding of each point-support node to a scalar field estimate with no output activation. Since rainfall is non-negative, predictions are clamped to be non-negative at evaluation time rather than constrained during training; we found that avoiding an output activation during training prevents vanishing gradients across many dry periods in the record. Because a distinct weight matrix $\mathbf{W}_r^{(l)}$ is learned per edge type, the model can weight each support relation differently rather than treating all incoming observations identically, while the shared aggregation keeps the architecture invariant to the number and composition of support layers present.

\subsection{Inductive Training and Field Reconstruction}
\label{sec:training}

\paragraph{Leave-one-out masked training.} 
A difficulty specific to this setting is that point-support observations serve as both input features and prediction targets. Without intervention, the model can learn a trivial identity mapping that copies each node's own observation to its output, minimizing training loss without learning any spatial relationship from its neighbors. To prevent this, we adopt a leave-one-out masking scheme, following the inductive mask-and-reconstruct principle used in graph-based kriging and imputation \citep{ignnk2021, marisca2022}. At each training step, a single point-support node has its observation masked, and the model predicts that node's value from its neighbors only. Crucially, only the observed value and validity flag are masked; the Laplacian positional encoding is preserved, so the masked node retains its structural identity in the graph while losing its observed value. The loss is computed solely on the masked node, 
\begin{equation}
    \mathcal{L} = \big( \hat{y}_v - y_v \big)^2,
    \label{eq:loss}
\end{equation}
where \(y_v\) is the observed rainfall at the masked point-support node \(v\), and \(\hat{y}_v\) is the corresponding model prediction. This masking-and-prediction step is repeated once for every point-support node in the graph, with gradients accumulated across all such passes before a single parameter update. No node is ever exposed to its own ground-truth value while learning to predict it. At evaluation time, all held-out nodes are masked simultaneously in a single pass, since their predictions are mutually independent given the observed nodes and carry no risk of leakage between one another. 

\textbf{Leakage control through graph reconstruction.}
The graph is rebuilt independently for each data split. Held-out nodes enter only the evaluation graph, never the training graph, so no information about their location or connectivity influences training. Normalisation statistics are likewise computed on the training split alone.

\textbf{Field reconstruction on user-defined target grids.}
Because the model predicts at a masked point-support node using only messages from its neighbors, the prediction is not tied to the locations observed during training. At inference, a regular grid of target nodes is inserted into the point-support layer. Each target node is connected to nearby observed point nodes through within-support edges and to nearby line and gridded observations through the same cross-support edge rules used during training. A single forward pass then yields a field estimate at every target node. The graph is therefore dynamic at inference: the resolution of the reconstructed field is determined by the density of inserted target nodes and is decoupled from the resolution of any input source. The same trained model can be queried on user-defined target grids without retraining.

\section{Experimental Setup}
\label{sec:experiments}
We evaluate the framework on rainfall estimation, a problem that is a natural instantiation of multi-support fusion. Rainfall is one of the few physical fields observed simultaneously through all three support types: rain gauges provide accurate point (0D) measurements at fixed locations; CMLs provide path-integrated (1D) observations of attenuation along communication paths; and weather radar and satellite products provide gridded (2D) estimates over regular cells. These sources are individually limited and mutually complementary: gauges are precise but spatially sparse, gridded products offer broad coverage but are biased and indirect, and links offer near-surface path-averaged information between the two, which makes their fusion a demanding test of whether geometry-aware multi-support modeling improves the reconstruction of a single underlying field.

\subsection{Datasets}
\label{sec:datasets}

We instantiate the framework on two geographically and climatically distinct testbeds with different support compositions, demonstrating that the same architecture applies without modification across sensor configurations. 

\paragraph{Singapore (0D + 1D + 2D).}
The Singapore testbed combines all three support types over the period 17 May 2024 to 30 April 2025. It comprises 70 rain gauge stations (0D) regulated by the National Environment Agency \citep{nea_rainfall_2024}, 414 CMLs (1D), and weather radar (2D) at $1$\,km resolution, the latter two provided by Pluvia.ai \citep{pluvia_2025}. The radar product spans a $240 \times 240$\,km domain; evaluation is performed at the gauge stations over the Singapore landmass (approximately $50 \times 27$\,km).

\paragraph{Sydney (0D + 2D + 2D).}
The Sydney testbed combines point and two distinct gridded supports over the Greater Sydney region (approximately $70 \times 55$\,km) for 2022 at hourly resolution. It comprises 75 gauge stations (0D) \citep{bom_waterdata}, gauge-corrected weather radar (2D) from the operational Rainfields\,3 product \citep{bom_rainfields3_2022} subsampled to approximately $5$\,km spacing, and satellite retrievals (2D) from the Himawari-8/9 CRRPH product \citep{bom_crrph_2024} at $2$\,km resolution. This composition exercises the framework's ability to fuse multiple heterogeneous gridded sources within the 2D support layer, each represented as a distinct node type sharing the same cross-support construction.

\subsection{Evaluation Protocol}
\label{sec:eval_protocol}

We evaluate every method under five-fold spatial cross-validation, in which held-out gauge stations are masked and predicted from the remaining observations. Stations are partitioned into spatially coherent folds so that held-out stations are not adjacent to training stations, preventing a model from succeeding through trivial spatial proximity rather than genuine generalisation. The same held-out stations are used across all methods and all support configurations to ensure comparability. Normalisation statistics are computed on the training split alone and applied to the held-out stations. 
Predictions are evaluated against the held-out gauge observations using RMSE and Pearson correlation ($r$), reported as the mean across folds. 

\subsection{Implementation Details}

Singapore experiments use 15-minute data, with gauge readings resampled from 5-minute accumulations to align with CML and radar timestamps; Sydney experiments use hourly data. Graphs are built with $k$-nearest-neighbor search over sensor coordinates, using $k=5$ for point-point and CML-point connections and $k=9$ for grid-point connections. Features are z-score normalized per node type using training-split statistics only. The HGNN uses five message-passing layers, hidden dimension 32, and dropout $p=0.15$. All neural models are trained with Adam ($3{\times}10^{-4}$ learning rate, $10^{-4}$ weight decay), weighted MSE loss, and early stopping with patience 10.

\subsection{Baselines}
\label{sec:baselines}

We compare against three tiers of baseline, each isolating a different aspect of the contribution. 

\textbf{Classical interpolation.}
IDW represents the geometry-agnostic, non-learned interpolation methods in operational use, and establishes the performance floor that any learned method must exceed. Among classical statistical interpolators, geostatistical kriging variants such as ordinary kriging and kriging with external drift incorporate spatial covariance structure and auxiliary predictors such as elevation \citep{goovaerts2000, Merging_review}. In preliminary evaluation on our data both variants underperformed IDW, consistent with the known degradation of kriging in sparse, non-stationary regimes \citep{Merging_review}; we therefore adopt IDW as the best-performing classical baseline. 

\textbf{Learned non-graph fusion.} Convolutional neural networks represent learned fusion without graph structure, operating on gridded inputs rather than an explicit graph. We adapt the spatiotemporal fusion architecture of \citet{wu2020deepfusion}, which extracts spatial features from co-located gridded sources through per-source convolutional branches and predicts rainfall at gauge locations. The inputs are three gridded channels  (an inverse-distance interpolated gauge field, gridded CML, and gridded radar), and the model is trained and evaluated under the same spatial cross-validation protocol as the graph models. We report two variants from this family: a single-frame convolutional model (CNN) that fuses the gridded sources at a single time step, and a convolutional-recurrent model (CNN-LSTM) that adds an LSTM to capture temporal dependence across consecutive frames. Together these isolate the value of graph structure: both see the same observations as our model but represent them on a fixed grid rather than as geometry-aware support-typed nodes.

\textbf{Support-agnostic HGNN.}
The most direct ablation of the contribution. This baseline uses an identical architecture, node features, capacity, training scheme, and data, but represents all observations as generic point nodes: gridded cells and link endpoints are inserted as ordinary 0D nodes connected by plain distance-based edges, with a single shared convolution in place of support-typed message passing. It retains the same observational information as our model but discards the support geometry. Improvements over this baseline isolate the effect of support-aware construction as much as possible while holding model capacity and inputs fixed.

\section{Results and Discussion}
\label{sec:results}

\subsection{Multi-Support Fusion on the Singapore Testbed}
\label{sec:results_sg}

\begin{table}[htbp]
\centering
\caption{Reconstruction performance at held-out gauge locations on the Singapore testbed (0D, 1D, and 2D supports). RMSE in mm\,h$^{-1}$ and Pearson $r$ against withheld gauge values; $\Delta\%$ is relative change versus the IDW baseline. The convolutional baseline consumes the sources as gridded channels. Supports: G (gauge), C (CML), R (radar). Evaluation follows the spatial cross-validation protocol of
Section~\ref{sec:eval_protocol}. Best in bold.}
\label{tab:singapore}
\setlength{\tabcolsep}{4pt}
\renewcommand{\arraystretch}{1.1}
\begin{tabular}{@{}lcccc@{}}
\toprule
Model & RMSE $\downarrow$ & \shortstack{$\Delta\%$\\vs IDW} & \shortstack{Pearson\\$r$ $\uparrow$} & \shortstack{$\Delta\%$\\vs IDW} \\
\midrule
IDW (baseline) & 2.429 & ---     & 0.714 & ---     \\
CNN (G+C+R)    & 2.254 & $-7.2$  & 0.697 & $-2.4$  \\
CNN-LSTM (G+C+R) & 2.047 & $-15.7$ & 0.744 & $+4.2$ \\
HGNN (sup. agnostic) & 2.130 & $-12.3$ & 0.737 & $+3.2$ \\
\midrule
HGNN (G)       & 1.946 & $-19.9$ & 0.780 & $+9.2$  \\
HGNN (G+C)     & 1.927 & $-20.7$ & 0.789 & $+10.4$ \\
HGNN (G+C+R)   & \textbf{1.866} & \textbf{--23.2} & \textbf{0.793} & \textbf{+11.1} \\
\bottomrule
\end{tabular}
\end{table}

Table~\ref{tab:singapore} reports reconstruction performance at held-out gauge locations on the Singapore testbed, where 0D (gauge), 1D (CML), and 2D (radar) supports are available. Each model predicts at gauge locations held out during evaluation and is scored against the withheld observations using RMSE (mm\,h$^{-1}$) and Pearson correlation, with relative change against the IDW baseline. 

\begin{figure}[t]
  \centering
  \includegraphics[width=0.5\textwidth]{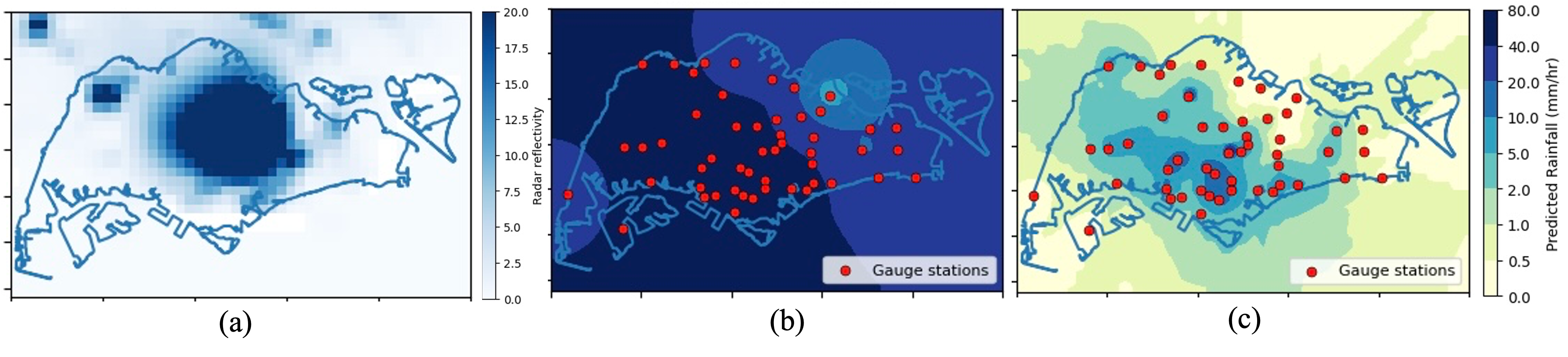}
  \caption{(a) Radar reflectivity (spatial reference only), (b) gauge-only IDW, and (c) multi-support HGNN
(gauge + CML + radar) prediction at 2025-04-20 17:00, Singapore.}
  \label{fig:qualitative_structure}
\end{figure}

The following findings stand out. First, the learned graph model improves substantially over classical interpolation even using gauges alone: the gauge-only HGNN reduces RMSE by $19.9\%$ (1.946 versus 2.429) and raises correlation by $9.2\%$ (0.780 versus 0.714) relative to IDW. 

Second, performance improves monotonically as supports of differing geometry are added: introducing the 1D CML support lowers RMSE to 1.927, and adding the 2D radar support lowers it further to 1.866, a $23.2\%$ reduction over IDW and the best result in the table. This progression is the central evidence for the multi-support hypothesis: each additional support type, with its own geometry encoded in the graph, contributes information that the others do not. 

Third, the most telling finding for the framing of this work, the single-frame convolutional baseline (CNN) that consumes the same gridded inputs as stacked image channels attains an RMSE of only 2.254, a mere $7.2\%$ improvement over IDW and worse than even the gauge-only HGNN. Adding temporal modeling (CNN-LSTM) helps, lowering RMSE to 2.047 and lifting correlation above IDW, but it still trails every HGNN variant, including the gauge-only model that uses strictly less information. The gridded sources carry the same raw information in both the convolutional and graph models; the difference is that the CNN treats them as dense image channels registered to a pixel grid, discarding their geometric relationship as 2D supports to the underlying field and to the off-grid 0D and 1D observations. The support-aware graph, which preserves that geometry, extracts markedly more from the identical inputs. %Treating heterogeneous observations as images is therefore not sufficient: it is the representation of measurement support, not the raw availability of the sources, that drives the improvement. 

Figure~\ref{fig:qualitative_structure} illustrates this mechanism at a single convective timestep. Radar is shown as a spatial structural reference only; its magnitude is biased relative to gauge observations and the panels use independent scales. Radar resolves a compact, off-gauge cell that gauge-only IDW cannot recover, smoothing the field into broad structure around the gauge locations; the multi-support HGNN, fusing the same radar with gauges and links, reconstructs the cell while remaining calibrated to gauge magnitudes. Radar magnitude is biased but is accepted as a reliable indicator of spatial structure \citep{Merging_review}, and fusing it recovers structure that gauge-only interpolation cannot. The reconstructed fields shown are produced at a chosen display resolution by the same trained model, by inserting target nodes on a regular grid (Section~\ref{sec:method}); the resolution is a free parameter at inference rather than a property of any input source. 

\paragraph{Support-agnostic baseline.} 
Fourth, the gain is largely associated with support-aware construction rather than graph structure alone. A support-agnostic HGNN, identical in architecture, features, capacity, and training but representing every observation as a 0D node connected by plain distance-based edges, attains an RMSE of 2.130, recovering only part of the gap between IDW and the full multi-support model. Stripping the support typing while holding everything else fixed forfeits roughly half of the improvement the support-typed model achieves over IDW (a $12.3\%$ reduction versus $23.2\%$). Where the convolutional baseline shows that images discard geometry, this baseline shows that a graph alone is not sufficient: the same observations, given a graph but stripped of their support typing, leave much of the available skill unrecovered. This comparison indicates that much of the gain comes from support-aware construction rather than graph structure alone.

% \begin{figure*}
%   \centering
%   \begin{subfigure}{0.68\linewidth}
%     \fbox{\rule{0pt}{2in} \rule{.9\linewidth}{0pt}}
%     \caption{An example of a subfigure.}
%     \label{fig:short-a}
%   \end{subfigure}
%   \hfill
%   \begin{subfigure}{0.28\linewidth}
%     \fbox{\rule{0pt}{2in} \rule{.9\linewidth}{0pt}}
%     \caption{Another example of a subfigure.}
%     \label{fig:short-b}
%   \end{subfigure}
%   \caption{Example of a short caption, which should be centered.}
%   \label{fig:short}
% \end{figure*}

% \begin{figure*}
% \centering  
% \includegraphics[width=1\textwidth]{radar_event.png}\\[0.5em]
% \includegraphics[width=1\textwidth]{idw_timeseries.png}
% \includegraphics[width=1\textwidth]{raingauge_timeseries.png}\\[0.5em]
% \includegraphics[width=1\textwidth]{raingauge_cml_timeseries.png} \\ [0.5em]
% \includegraphics[width=1\textwidth]{raingauge_cml_radar_timeseries.png} \\ [0.5em]
% \caption{Radar image(top), IDW(2nd row), Gauge(3rd row), Gauge+CML(4th), Gauge-CML-Radar(Bottom)}
% \label{fig:stacked_timeseries}
% \end{figure*}
% \FloatBarrier

\subsection{Generalization to the Sydney Testbed}

\begin{table}[htbp]
\centering
\caption{Generalization to the Sydney testbed (point and two gridded
supports). RMSE in mm\,h$^{-1}$ and Pearson $r$ against withheld gauge
values; $\Delta\%$ is relative change versus the IDW baseline. Evaluation
follows the protocol of Section~\ref{sec:eval_protocol}. Supports:
G (gauge), R (radar), S (satellite). Best in bold.}
\label{tab:sydney}
\setlength{\tabcolsep}{4pt}
\renewcommand{\arraystretch}{1.1}
\begin{tabular}{@{}lcccc@{}}
\toprule
Model & RMSE $\downarrow$ & \shortstack{$\Delta\%$\\vs IDW} & \shortstack{Pearson\\$r$ $\uparrow$} & \shortstack{$\Delta\%$\\vs IDW} \\
\midrule
IDW (baseline)  & \textbf{0.667} & ---    & \textbf{0.854} & ---    \\
HGNN (G)        & 0.686 & $+2.9$ & 0.843 & $-1.3$ \\
HGNN (G+R)      & 0.688 & $+3.2$ & 0.842 & $-1.4$ \\
HGNN (G+R+S)    & 0.698 & $+4.8$ & 0.840 & $-1.6$ \\
\bottomrule
\end{tabular}
\end{table}
The Singapore results show large and consistent gains. To probe the boundary of where these gains hold, we apply the same framework to a Sydney testbed with different support configuration (0D and two 2D sources, radar and satellite; no 1D support) and a different rainfall regime, evaluated at held-out gauge locations under the same spatial cross-validation protocol. Table~\ref{tab:sydney} reports the outcome: the gauge-only HGNN does not improve over IDW (correlation 0.843 versus 0.854; RMSE 0.686 versus 0.667), and adding the 2D radar and satellite supports does not change this picture (correlation 0.842 with radar, 0.840 with radar and satellite). 

Rather than treat this as an isolated negative result, we find that a single mechanism explains both the Singapore success and the Sydney plateau: the skill available to a learned or fused estimator appears to depend on the gauge spacing \emph{relative to the spatial correlation length of the rainfall field}, not by gauge count. When a field is sampled finely relative to its correlation length, it is already well determined by the gauges; little residual structure remains for either a learned model or an additional gridded sensor to recover. When a fast-decorrelating field is sampled coarsely relative to its correlation length, substantial sub-gauge structure remains, and both learning and fusion have headroom.

The two testbeds sit on opposite sides of this divide, and the baseline correlations reveal it directly. Despite a \emph{denser} gauge network, Singapore's IDW baseline is markedly weaker than Sydney's (correlation 0.714 versus 0.854): interpolation skill appears to depend on the field's spatial correlation, not the gauge count. Sydney's rainfall remains coherent across its gauge spacing: empirical inter-gauge Pearson correlation (computed from pairwise gauge correlations on the full gauge record) is approximately 0.85 at the $\sim$3\,km median spacing with a length near 14\,km, so a held-out location is already well predicted by its neighbours, leaving little for the model to add and IDW close to optimal. Singapore's convective rainfall, by contrast, decorrelates over a much shorter distance, leaving sub-gauge structure that fixed interpolation cannot resolve and that the support-aware model recovers. 

%The same mechanism explains why the 2D supports do not improve reconstruction in Sydney and slightly degrade the estimate. A prediction at a held-out point draws on its neighboring gauges and on the gridded radar or satellite pixel covering it. Because Sydney's field is coherent across its gauge spacing, the gauge-based estimate already carried low residual uncertainty, while the radar and satellite products are coarser, biased observations of the same quantity; adding a noisier estimate of an already well-estimated value does not reduce the error and, as Table~\ref{tab:sydney} shows, increases it monotonically as lower-quality sources are added, with  the satellite addition producing the largest degradation. The radar support, by contrast, improves reconstruction in Singapore precisely because the under-sampled convective field there leaves residual uncertainty for a direct observation of the local cell to reduce. The same class of 2D gridded observation is therefore informative or harmful depending not on the sensor but on whether the field retains sub-gauge structure for it to resolve. 

This also explains why adding Sydney’s gridded products does not improve the estimate as Table~\ref{tab:sydney} shows. When nearby gauges already provide a low-uncertainty estimate, coarser radar or satellite observations may add bias rather than useful residual structure. In Singapore, by contrast, the under-sampled convective field leaves sub-gauge structure that radar can help localize.

We note two honest limitations of this argument. The Singapore and Sydney testbeds differ along two axes simultaneously (Singapore additionally possesses a 1D (CML) support and exhibits faster decorrelation) and our data cannot fully separate the two; we therefore describe the correlation-length account as consistent with the evidence rather than proven. The gauge-only comparison, in which no 1D support is involved, provides the cleaner support for the mechanism, since there the entire gap must be attributed to the field's spatial correlation rather than to support complementarity. Taken together, the two testbeds delineate the operating envelope of multi-support fusion: it delivers the largest gains where supports are geometrically complementary and the field is under-sampled relative to its correlation length, and offers little where the field is already well resolved by existing 0D observations.
\section{Conclusion}
We introduced a geometry-aware multi-support HGNN that represents observations according to their spatial support, point, line or gridded, and fuses them through cross-support message passing into a shared prediction layer. By making the structural distinction between 0D, 1D, and 2D supports explicit in the graph rather than absorbing it into feature space, the framework treats each measurement as the geometric constraint it actually imposes on the field. An inductive masked-node formulation lets a single trained model reconstruct the field at user-defined target grids, decoupled from the resolution of any input source. 

On rainfall estimation, encoding support geometry improved reconstruction RMSE by 23.2\% over classical interpolation and outperformed both the convolutional fusion baseline and a support-agnostic graph of matched capacity, indicating that support-aware construction, rather than graph structure alone, accounts for much of the gain. A second testbed with a different support composition showed where the approach helps and where it does not: multi-support fusion delivers the largest gains when the field is under-sampled relative to its spatial correlation length, and little when existing point observations already resolve the field. This account of when fusion helps, rather than a single benchmark number, is what we expect to carry over to other multi-support estimation problems. 

The framework is limited to the support types instantiated here, and the two testbeds differ along more than one axis, so the correlation-length account is consistent with our evidence rather than proven. Future work will evaluate additional cities, richer line- and area-support encodings, temporal forecasting, and downstream hydrological impact.

{
    \small
    \bibliographystyle{ieeenat_fullname}
    \bibliography{main}

@article{Merging_review,
author = {Ochoa-Rodriguez, Susana and Wang, Li-Pen and Willems, Patrick and Onof, Christian},
year = {2019},
month = {08},
pages = {},
title = {A Review of Radar‐Rain Gauge Data Merging Methods and Their Potential for Urban Hydrological Applications},
volume = {55},
journal = {Water Resources Research},
doi = {10.1029/2018WR023332}
}

@article{Bianchi,
author = {Bianchi, Blandine and Van Leeuwen, Peter Jan and Hogan, Robin and Berne, Alexis},
year = {2013},
month = {12},
pages = {1897-1909},
title = {A Variational Approach to Retrieve Rain Rate by Combining Information from Rain Gauges, Radars, and Microwave Links},
volume = {14},
journal = {Journal of Hydrometeorology},
doi = {10.1175/JHM-D-12-094.1}
}

@article{CNGAT,
author = {Peng, Xuan and Li, Qian and Jing, Jinrui},
year = {2021},
month = {10},
pages = {1-1},
title = {CNGAT: A Graph Neural Network Model for Radar Quantitative Precipitation Estimation},
volume = {PP},
journal = {IEEE Transactions on Geoscience and Remote Sensing},
doi = {10.1109/TGRS.2021.3120218}
}

@article{acosta2025physics,
  title={Physics-informed Graph Neural Networks for Operational Flood Modeling},
  author={Carlo Malapad Acosta and Herath Mudiyanselage Viraj Vidura Herath and Jia Yu Lim and Abhishek Saha and Sanka Rasnayaka and Lucy Amanda Marshall},
  journal={ArXiv},
  year={2025},
  volume={abs/2512.23964},
  url={https://api.semanticscholar.org/CorpusID:288258177}
}

@article{HGNNSurvey,
  title={A Survey on Heterogeneous Graph Embedding: Methods, Techniques, Applications and Sources},
  author={Xiao Wang and Deyu Bo and Chuan Shi and Shaohua Fan and Yanfang Ye and Philip S. Yu},
  journal={IEEE Transactions on Big Data},
  year={2020},
  volume={9},
  pages={415-436},
  url={https://api.semanticscholar.org/CorpusID:227229005}
}

@article{lam2023graphcastlearningskillfulmediumrange,
author = {Lam, Remi and Sanchez-Gonzalez, Alvaro and Willson, Matthew and Wirnsberger, Peter and Fortunato, Meire and Alet, Ferran and Ravuri, Suman and Ewalds, Timo and Eaton-Rosen, Zach and Hu, Weihua and Merose, Alexander and Hoyer, Stephan and Holland, George and Vinyals, Oriol and Stott, Jacklynn and Pritzel, Alexander and Mohamed, Shakir and Battaglia, Peter},
year = {2023},
month = {11},
pages = {eadi2336},
title = {Learning skillful medium-range global weather forecasting},
volume = {382},
journal = {Science (New York, N.Y.)},
doi = {10.1126/science.adi2336}
}

@article{yang2025offgrid,
author = {Yang, Qidong and Giezendanner, Jonathan and Civitarese, Daniel Salles and Jakubik, Johannes and Schmitt, Eric and Chandra, Anirban and Vila, Jeremy and Hohl, Detlef and Hill, Chris and Watson, Campbell and Wang, Sherrie},
title = {Local Off-Grid Weather Forecasting With Multi-Modal Earth Observation Data},
journal = {Journal of Advances in Modeling Earth Systems},
volume = {17},
number = {12},
pages = {e2025MS005207},
doi = {https://doi.org/10.1029/2025MS005207},
url = {https://agupubs.onlinelibrary.wiley.com/doi/abs/10.1029/2025MS005207},
eprint = {https://agupubs.onlinelibrary.wiley.com/doi/pdf/10.1029/2025MS005207},
note = {e2025MS005207 2025MS005207},
year = {2025}
}

@article{Rainlink,
author = {Overeem, A. and Leijnse, H. and Uijlenhoet, Remko},
year = {2015},
month = {08},
pages = {8191-8230},
title = {Retrieval algorithm for rainfall mapping from microwave links in a cellular communication network},
volume = {8},
journal = {Atmospheric Measurement Techniques Discussions},
doi = {10.5194/amtd-8-8191-2015}
}

@article{CMLReview,
author = {Chwala, Christian and Kunstmann, Harald},
year = {2019},
month = {03},
pages = {},
title = {Commercial microwave link networks for rainfall observation: Assessment of the current status and future challenges},
volume = {6},
journal = {Wiley Interdisciplinary Reviews: Water},
doi = {10.1002/wat2.1337}
}

@article{wu2020deepfusion,
author = {Wu, Hongcai and Yang, Qinli and Liu, Jiaming and Wang, Guoqing},
year = {2020},
month = {02},
pages = {124664},
title = {A spatiotemporal deep fusion model for merging satellite and gauge precipitation in China},
volume = {584},
journal = {Journal of Hydrology},
doi = {10.1016/j.jhydrol.2020.124664}
}

@article{CNNDataFusion,
author = {Moraux, Arthur and Dewitte, Steven and Cornelis, Bruno and Munteanu, Adrian},
year = {2021},
month = {08},
pages = {3278},
title = {A Deep Learning Multimodal Method for Precipitation Estimation},
volume = {13},
journal = {Remote Sensing},
doi = {10.3390/rs13163278}
}

@misc{nea_rainfall_2024,
  author       = {{National Environment Agency}},
  title        = {Rainfall Across {Singapore}},
  howpublished = {data.gov.sg, Dataset},
  year         = {2024},
  note         = {Retrieved April 8, 2026}
}

@misc{pluvia_2025,
  author       = {{Pluvia.ai}},
  title        = {Pluvia -- Commercial Microwave Link and Radar Data},
  howpublished = {CML and gauge-corrected radar data for {Singapore}, provided via research collaboration},
  year         = {2025}
}

@misc{bom_waterdata,
  author       = {{Bureau of Meteorology}},
  title        = {{WaterData}: Water Data Online},
  howpublished = {\url{http://www.bom.gov.au/waterdata/}},
  note         = {Retrieved February 2026}
}

@misc{bom_rainfields3_2022,
  author       = {{Bureau of Meteorology}},
  title        = {{AURA} -- Operational Radar {Rainfields\,3}},
  year         = {2022},
  note         = {Accessed via NCI project rq0}
}

@misc{bom_crrph_2024,
  author       = {{Bureau of Meteorology}},
  title        = {{Himawari} 8/9 Convective Rain Rate -- Physically Based ({CRRPH})},
  year         = {2024},
  note         = {Accessed via NCI project rv74}
}

@article{rfmep2020,
author = {Baez-Villanueva, Oscar and Zambrano-Bigiarini, Mauricio and Beck, Hylke and McNamara, Ian and Ribbe, Lars and Nauditt, Alexandra and Birkel, Christian and Verbist, Koen and Giraldo-Osorio, Juan Diego and Thinh, Nguyen},
year = {2020},
month = {03},
pages = {},
title = {RF-MEP: A novel Random Forest method for merging gridded precipitation products and ground-based measurements},
volume = {239},
journal = {Remote Sensing of Environment},
doi = {10.1016/j.rse.2019.111606}
}

@unknown{ignnk2021,
author = {Wu, Yuankai and Zhuang, Dingyi and Labbe, Aurelie and Sun, Lijun},
year = {2020},
month = {06},
pages = {},
title = {Inductive Graph Neural Networks for Spatiotemporal Kriging},
doi = {10.48550/arXiv.2006.07527}
}

@article{dwivedi2021graphtransformer,
  author       = {Vijay Prakash Dwivedi and Xavier Bresson},
  title = {A Generalization of Transformer Networks to Graphs},
  journal = {CoRR},
  volume = {abs/2012.09699},
  year = {2020},
  url = {https://arxiv.org/abs/2012.09699},
  eprinttype   = {arXiv},
  eprint       = {2012.09699},
  bibsource    = {dblp computer science bibliography, https://dblp.org}
}

@article{pyg2019,
  author       = {Matthias Fey and
                  Jan Eric Lenssen},
  title        = {Fast Graph Representation Learning with PyTorch Geometric},
  journal      = {CoRR},
  volume       = {abs/1903.02428},
  year         = {2019},
  url          = {http://arxiv.org/abs/1903.02428},
  eprinttype   = {arXiv},
  eprint       = {1903.02428},
  bibsource    = {dblp computer science bibliography, https://dblp.org}
}

@article{goovaerts2000,
author = {Goovaerts, Pierre},
year = {2000},
month = {02},
pages = {113-129},
title = {Geostatistical Approaches for Incorporating Elevation Into the Spatial Interpolation of Rainfall},
volume = {228},
journal = {Journal of Hydrology},
doi = {10.1016/S0022-1694(00)00144-X}
}

@misc{marisca2022,
      title={Learning to Reconstruct Missing Data from Spatiotemporal Graphs with Sparse Observations}, 
      author={Ivan Marisca and Andrea Cini and Cesare Alippi},
      year={2022},
      eprint={2205.13479},
      archivePrefix={arXiv},
      primaryClass={cs.LG},
      url={https://arxiv.org/abs/2205.13479}, 
}

@article{AdvancesInMSDF,
author = {Chen, Hanqing and Zeng, Jiangyuan and Lyu, Yi and Yong, Bin},
year = {2026},
month = {01},
pages = {100195},
title = {Multisource precipitation data fusion: Generating high-quality precipitation estimates},
volume = {4},
journal = {The Innovation Geoscience},
doi = {10.59717/j.xinn-geo.2026.100195}
}
}

\end{document}